\documentclass[conference]{IEEEtran}
\IEEEoverridecommandlockouts

\usepackage{cite}
\usepackage{amsmath,amssymb,amsfonts}
\usepackage{algorithmic}
\usepackage{graphicx}
\usepackage{textcomp}
\usepackage{xcolor}
\usepackage{graphicx}
\usepackage{amsmath}
\usepackage{multirow}
\usepackage{xcolor,colortbl}
\definecolor{lightgrey}{rgb}{0.93,0.93,0.93}
\usepackage[linesnumbered,ruled,vlined]{algorithm2e}
\usepackage{subcaption}
\usepackage{bbding}
\usepackage[normalem]{ulem}
\usepackage{enumitem}
\usepackage{url}
\usepackage{hyperref}
\usepackage[numbers]{natbib} 
\usepackage{booktabs}
\usepackage{tcolorbox}
\usepackage{tabularx}
\usepackage[dvipsnames]{xcolor}
\usepackage{pifont}
\usepackage[table]{xcolor}
\tcbuselibrary{breakable}

\usepackage[framemethod=tikz]{mdframed}
\mdfdefinestyle{customstyle}{
  linecolor=gray!100,
  linewidth=3pt,
  innerleftmargin=3pt,
  topline=false,
  rightline=false,
  bottomline=false,
  leftline=true,
  innerrightmargin=3pt,
  innertopmargin=3pt,
  innerbottommargin=3pt,
  backgroundcolor=gray!15
}

\newenvironment{custommdframed}
  {\begin{mdframed}[style=customstyle]}
  {\end{mdframed}}
  
\definecolor{c4}{cmyk}{0.6765,0.2017,0,0.0667}
\newtcbox{\hlprimarytab}{on line, rounded corners, box align=base, colback=c3!10,colframe=white,size=fbox,arc=3pt, before upper=\strut, top=-2pt, bottom=-4pt, left=-2pt, right=-2pt, boxrule=0pt}
\newtcbox{\hlsecondarytab}{on line, box align=base, colback=red!10,colframe=white,size=fbox,arc=3pt, before upper=\strut, top=-2pt, bottom=-4pt, left=-2pt, right=-2pt, boxrule=0pt}
\newtcbox{\hlthirdtab}{on line, rounded corners, box align=base, colback=c4!10,colframe=white,size=fbox,arc=3pt, before upper=\strut, top=-2pt, bottom=-4pt, left=-2pt, right=-2pt, boxrule=0pt, fontupper=\scriptsize}

\newcommand{\comparison}[1]{{\small\hlthirdtab{#1}}}

\newcommand{\ourmodel}{\textsc{FastCoder}}

\def\BibTeX{{\rm B\kern-.05em{\sc i\kern-.025em b}\kern-.08em
    T\kern-.1667em\lower.7ex\hbox{E}\kern-.125emX}}
\begin{document}

\title{\ourmodel{}: Accelerating Repository-level Code Generation via Efficient Retrieval and Verification}

\author{
    \IEEEauthorblockN{Qianhui Zhao$^{1}$, Li Zhang$^{1}$, Fang Liu$^{1*}$\thanks{* Corresponding author}, Xiaoli Lian$^{1*}$, Qiaoyuanhe Meng$^{1}$  \\
    Ziqian Jiao$^{1}$, Zetong Zhou$^{1}$, Jia Li$^{3}$, Lin Shi$^{2}$}
    \IEEEauthorblockA{$^{1}$State Key Laboratory of Complex \& Critical Software Environment, School of Computer Science and Engineering \\ Beihang University, China}
    \IEEEauthorblockA{$^{2}$School of Software, Beihang University, China}
    \IEEEauthorblockA{$^{3}$Peking University, China}
    \IEEEauthorblockA{
    Email: \{\href{mailto:zhaoqianhui@buaa.edu.cn}{zhaoqianhui}, \href{mailto:fangliu@buaa.edu.cn}{fangliu}, \href{mailto:fangliu@buaa.edu.cn}{lianxiaoli}\}@buaa.edu.cn
    }
}

\maketitle

\begin{abstract}
Code generation is a latency-sensitive task that demands high timeliness. However, with the growing interest and inherent difficulty in repository-level code generation, most existing code generation studies focus on improving the correctness of generated code while overlooking the inference efficiency, which is substantially affected by the overhead during LLM generation. Although there has been work on accelerating LLM inference, these approaches are not tailored to the specific characteristics of code generation; instead, they treat code the same as natural language sequences and ignore its unique syntax and semantic characteristics, which are also crucial for improving efficiency. Consequently, these approaches exhibit limited effectiveness in code generation tasks, particularly for repository-level scenarios with considerable complexity and difficulty. To alleviate this issue, following draft-verification paradigm, we propose \ourmodel{}, a simple yet highly efficient inference acceleration approach specifically designed for code generation, without compromising the quality of the output. \ourmodel{} constructs a multi-source datastore, providing access to both general and project-specific knowledge, facilitating the retrieval of high-quality draft sequences. Moreover, \ourmodel{} reduces the retrieval cost by controlling retrieval timing, and enhances efficiency through parallel retrieval and a context- and LLM preference-aware cache. Experimental results show that \ourmodel{} can reach up to $2.53 \times$ and $2.54\times$ speedup compared to autoregressive decoding in repository-level and standalone code generation tasks, respectively, outperforming state-of-the-art inference acceleration approaches by up to $88\%$. \ourmodel{} can also be integrated with existing correctness-focused code generation approaches to accelerate the LLM generation process, and reach a speedup exceeding $2.6 \times$.

\end{abstract}


\section{Introduction}
Code generation is a crucial task in software engineering, which can reduce manual effort by automating repetitive and time-consuming programming tasks.
Large Language Models (LLMs), such as DeepSeek-Coder \cite{guo2024deepseek}, CodeLlama \cite{roziere2023codellama}, GPT-4o \cite{achiam2023gpt4}, \textit{etc.}, have demonstrated impressive performance in code generation tasks, revolutionizing the landscape of software development \cite{GitHub-Copilot, li2023starcoder}. 
Substantial researches have been proposed to improve the correctness of code (evaluated by the passing rate \cite{chen2021codex}) generated by LLMs \cite{liang2024repofuse, phan2024repohyper, liu2024graphcoder}. 
While code correctness ensures the program performs its intended behaviors accurately, efficiency of code generation is also crucial for real-world software development but often overlooked.
Therefore, optimizing the efficiency of code generation approaches has become a pressing issue.

In real-world software development, developers typically work within a specific context—such as a repository—that contains rich domain knowledge.
Since target code often relies on functions or classes already defined in the repository, effective repository-level code generation must consider the context of the entire repository. 
However, due to the large scale of code repositories and the limited context window of LLMs, it is infeasible to directly leverage the entire repository as input.
Thus, most existing studies adopt a retrieval-augmented generation (RAG) strategy \cite{cheng2024dataflow, zhang2023repocoder, liu2024graphcoder}, using the natural language requirement or partially written code as a query to retrieve relevant code snippets from the repository, which are then combined with the original query to guide code generation.

The RAG strategy can bring benefits to the functional correctness of the generated code, but neglects generation efficiency.
The latency of this paradigm is primarily composed of two stages: the retrieval stage and generation stage.
As we all know, during the generation stage, LLMs face the challenge of the significant inference time caused by autoregressive decoding mechanism. In this process, each new token is generated sequentially, conditioned on all previously generated tokens and the given context, resulting in repeated and computationally expensive forward passes.
Moreover, the growing complexity of relevant context retrieval and prompt construction in RAG approaches further increases the overall time cost of code generation. 
However, developers typically hold high expectations regarding the responsiveness of code recommendations \cite{liu2024non}. If LLMs fail to deliver precise and efficient feedback, it may directly affect development efficiency and user experience.

Although there exist approaches for LLM inference acceleration, they exhibit limited effectiveness in code generation, especially in repository-level code generation.
For example, Self-speculative decoding \cite{zhang2024draft} achieves approximately $1.5 \times$ acceleration compared to autoregressive decoding in standalone code generation (Fig. \ref{fig: humaneval results}), but falls short when applied to repository-level tasks, offering virtually no speedup (Table \ref{tab: main results}).
Most of these approaches follow a draft-verification paradigm \cite{zhang2024draft, zhao2024ouroboros, miao2024specinfer, he2024rest}, which utilizes a small model or a retrieval system to draft several candidate output tokens, and then employs the target LLM to verify the acceptability of draft tokens through a single forward step, keeping the output consistent with that decoded autoregressively by the target LLM itself.
The suboptimal performance on repository-level code generation may be attributed to the reason that existing approaches treat source code as sequences similar to natural language, without accounting for code's unique syntactic and semantic characteristics.
For instance, the fixed retrieval datastore of retrieval-based acceleration approach struggles to provide high-quality drafts for different repositories that follow customized functions and coding styles, and cannot adapt to output preferences of different LLMs. Additionally, performing retrieval uniformly across all code locations overlooks the varying difficulty levels of code generation at different positions.
Therefore, \textbf{there lack acceleration techniques specifically tailored for repository-level code generation tasks, which are orthogonal to correctness-focused approaches, enabling improvements in inference speed without compromising correctness performance.}

To alleviate this issue, in this paper, we primarily focus on improving the inference speed of LLMs on repository-level code generation task, without compromising the quality of the output. 
We propose \ourmodel{}, a \textbf{simple yet highly efficient} approach to accelerate the inference of LLMs through an efficient and effective retrieval strategy.
Concretely, to better align with the characteristics of different repositories, we first construct a multi-source datastore, providing access to both general and project-specific knowledge and enhancing the quality of draft sequences. 
Then, \ourmodel{} reduces unnecessary retrieval overhead by controlling the retrieval timing. Besides, \ourmodel{} improves retrieval efficiency through parallel retrieval and the maintenance of a context- and LLM preference-aware cache. Finally, draft sequences are constructed using a weighted Trie, and tree attention is employed to avoid redundant computation caused by verifying multiple draft sequences. 

Experimental results show that the decoding speed of \ourmodel{} surpasses existing inference acceleration approaches substantially on both repository-level and standalone code generation tasks. For repository-level code generation, \ourmodel{} achieves up to $2.30 \times$ and $2.53 \times$ speedup compared with autoregressive decoding on DevEval \cite{li2024deveval} and RepoEval \cite{zhang2023repocoder}, respectively.
\ourmodel{} can also achieve up to $2.54 \times$ acceleration on standalone code generation dataset, HumanEval \cite{chen2021codex}.
Moreover, it is worth mentioning that \ourmodel{} can be integrated with existing functional correctness-focused code generation methods to improve their generation efficiency while preserving the consistency of output sequences.
When combined with RepoCoder \cite{zhang2023repocoder} and RLCoder \cite{wang2024rlcoder}, \ourmodel{} can bring a speedup exceeding $2.6 \times$.

Our contributions can be summarized as follows:
\begin{itemize}[nosep]
    \item We identify limitations of current correctness-focused code generation studies and LLM inference acceleration approaches in repository-level code generation and provide insights for potential improvements.
    \item We propose \ourmodel{}, a simple yet efficient approach to accelerate LLM inference for code generation by leveraging effective retrieval and verification mechanisms, which can also be integrated with correctness-focused code generation approaches following RAG strategy.
    \item We conduct a comprehensive evaluation of the inference efficiency of \ourmodel{}, and results show that it achieves state-of-the-art results in both repository-level and standalone code generation tasks. We provide the code, data, and an initial demo at {\url{https://github.com/whisperzqh/FastCoder}}.
\end{itemize}

\section{Related Work} \label{sec: related work}

\subsection{Repository-level Code Generation}
LLMs have made significant progress in recent years, yet widely-used evaluation datasets for code generation task, such as HumanEval \cite{chen2021codex} and MBPP \cite{austin2021mbpp}, predominantly focus on standalone code generation tasks. This over-simplified setting falls short of representing the real-world software development scenario where repositories span multiple files with numerous cross-file dependencies.
In practice, a substantial portion of code relies on methods and properties defined in other files. These non-standalone functions constitute more than 70\% of the functions in popular open-source projects, making evaluations based solely on standalone functions insufficient for assessing model effectiveness \cite{yu2024codereval, Chen2025Deep}.

Consequently, repository-level code generation has gained increasing attention due to the inter-dependencies among code files, such as cross-module API calls and shared global snippets, which are crucial for context-aware code generation.
However, due to the vast size of code repositories and the limitations of LLMs' context length, it is impossible to leverage the entire repository directly as context.
Most prior studies adopt a retrieval-augmented generation (RAG) strategy, using the natural language requirement or unfinished code as a query to retrieve snippets from the repository, which are then combined with the query for code generation.
RepoFuse \cite{liang2024repofuse} employs BM25 \cite{robertson2009bm25} for lexical retrieval based on textual similarity with the unfinished code.
RepoHyper \cite{phan2024repohyper} adopts dense retrieval by encoding code into vectors to capture semantic similarity.
RepoCoder \cite{zhang2023repocoder} enhances retrieval iteratively using intermediate completions to better align context and generation.
RLCoder \cite{wang2024rlcoder} leverages reinforcement learning to train the retriever without labeled data.
As context retrieved based on text similarity may be insufficiently relevant to generation targets, Draco \cite{cheng2024dataflow} constructs a repo-specific context graph via dataflow analysis to precisely retrieve relevant background knowledge.
GraphCoder \cite{liu2024graphcoder} integrates graph-based retrieval with LLMs to combine general and repository-specific knowledge for code generation. 
However, these approaches focus solely on improving the functional correctness of the generated code, while neglecting generation efficiency.
The construction of information-rich prompts further increases the overall time cost of code generation.

Researchers have also introduced several benchmarks to evaluate code generation in realistic scenarios.
RepoEval \cite{zhang2023repocoder} focuses on repository-level code completion covering line, API invocation, and function body granularities.
CoderEval \cite{yu2024codereval} is a context-aware benchmark that categorizes code generation tasks into six levels based on external dependencies, ranging from standalone to project-level functions.
CrossCodeEval \cite{ding2024crosscodeeval} is a multilingual benchmark that emphasizes cross-file contextual understanding and employs static analysis to identify examples requiring such context.
DevEval \cite{li2024deveval} aligns with real-world repositories in terms of code and dependency distributions, and includes comprehensive metadata annotated manually by developers.

\subsection{LLM inference acceleration approaches} 

Autoregressive decoding generates tokens in a step-by-step manner and results in a slow and costly decoding process.
To accelerate decoding, previous non-autoregressive decoding approaches \cite{ghazvininejad2019maskpredict,liu2024non} were proposed to generate multiple tokens in parallel. However, this parallelism often comes at the cost of degraded model performance.
To address this issue, researchers have explored decoding acceleration methods that preserve model performance. Among these, draft-verification approaches \cite{chen2023accelerating,miao2024specinfer,he2024rest} have gained widespread adoption, which employ an alternative method to quickly produce candidate output tokens and then utilize the target LLM to verify them in a single forward pass.
These methods can be further categorized into generation-based and retrieval-based approaches, depending on how the initial draft is produced.

For generation-based approaches, the draft token can be generated either by a small model or the target LLM itself.
Speculative decoding \cite{chen2023accelerating, leviathan2023fast} minimizes the target LLM forward steps by using a smaller model for drafting and then employing the target LLM to verify the draft in a low-cost parallel manner. 
Based on this, Ouroboros \cite{zhao2024ouroboros} generates draft phrases to parallelize the drafting process and lengthen drafts.
Specinfer \cite{miao2024specinfer} uses many draft models obtained from distillation, quantization, and pruning to conduct speculations together.
However, identifying an appropriate draft model continues to pose significant challenges, as it must align with the vocabulary of the target LLM and achieve a delicate balance between keeping quick decoding speed and ensuring output quality.
Thus, researchers have investigated utilizing the target LLM itself to generate efficient draft sequences.
Blockwise Decoding \cite{stern2018blockwisedecoding} and Medusa \cite{cai2024medusa} introduce multiple heads to enable parallel generation of multiple tokens per step.
Lookahead decoding \cite{fu2024lookahead} uses an n-gram pool to cache the historical n-grams generated so far. 
Eagle \cite{li2024eagle} conducts the drafting process at the more structured feature level.
Self-speculative decoding \cite{zhang2024draft} employs the target LLM with selectively certain intermediate layers skipped as the draft model.
However, using the target LLM itself to generate draft tokens often requires modifications to the model architecture, necessitating additional training to achieve optimal performance.

To avoid additional training and identifying a suitable draft model, as well as the extra computational overhead introduced by draft models, more recently, researchers have explored retrieval-based approaches, which replace the draft model with a retrieval system \cite{he2024rest,yang2023llma}. 
By searching in a retrieval datastore to obtain candidate sequences, these approaches can easily be ported to any LLM and reduce computational overhead.
LLMA \cite{yang2023llma} is an inference-with-reference decoding mechanism by exploiting the overlap between the output and the reference of an LLM. It provides generic speedup through speculative retrieval and batched verification.
REST \cite{he2024rest} replaces the parametric draft model with a non-parametric retrieval datastore.
As many subsequences during generation likely appear in the datastore, it can frequently generate multiple correct tokens per step.

Our approach adopts the retrieval-based acceleration paradigm following REST \cite{he2024rest}, but has distinctive innovations, as it is specifically tailored to the characteristics of code generation. In particular, we design a multi-source datastore to reduce retrieval overhead while enhancing the relevance of retrieved results, exploit structural properties of code to avoid unnecessary retrieval in positions where it is unlikely to be beneficial, and incorporate a caching mechanism to ensure that retrieved drafts are better aligned with the preferences of the LLM and the characteristics of the target repository.

\section{Preliminaries}

\subsection{Retrieval-based Speculative Decoding}

Building upon the draft-verification framework introduced by speculative decoding \cite{chen2023accelerating, leviathan2023fast}, retrieval-based decoding acceleration approaches leverage a retrieval mechanism to generate draft tokens \cite{he2024rest,yang2023llma}, which can eliminate the challenge of selecting an appropriate draft model and avoid additional training costs.
A notable example is Retrieval-Based Speculative Decoding (REST) \cite{he2024rest}, which has proven to be effective in standalone function generation task \cite{chen2021codex}. Below is an explanation of how it works.
Pre-built from a code corpus, the datastore $D = \{(c_i, t_i)\}$ serves as the source for the draft token sequence, where $c_i$ represents a context and $t_i$ represents the corresponding continuation of $c_i$.
As an alternative to the draft model, the objective of retrieval is to identify the most likely continuations of the current context from the datastore $D$ using a suffix match \cite{manber1993suffix}.
Specifically, given a context $s = (x_1,...,x_t)$, it aims to find contexts in $D$ that match the longest suffix of $s$.
Starting from a pre-defined match length upper limit $n_{max}$ (measured in the number of tokens), for each suffix length $n$, it extracts the suffix of $s$ with $n$ tokens, denoted as $q$, and obtains all contexts $c_i$ that match $q$ as a suffix. If at least one context in $D$ matches $q$, the corresponding context continuation pairs are returned as the retrieval result $S$; otherwise, the match length $n$ is decreased by one to attempt matching a shorter suffix. 
Subsequently, the top-$k$ high-frequency prefixes in $S$ are selected as the draft sequences for later verification.
Inspired by REST, \ourmodel{} also incorporates a similar suffix-match-based retrieval algorithm, leveraging its advantages in time and memory efficiency.

\begin{figure}[t]
 \setlength{\abovecaptionskip}{0.1cm}
    \centering
    \includegraphics[width=0.9\linewidth]{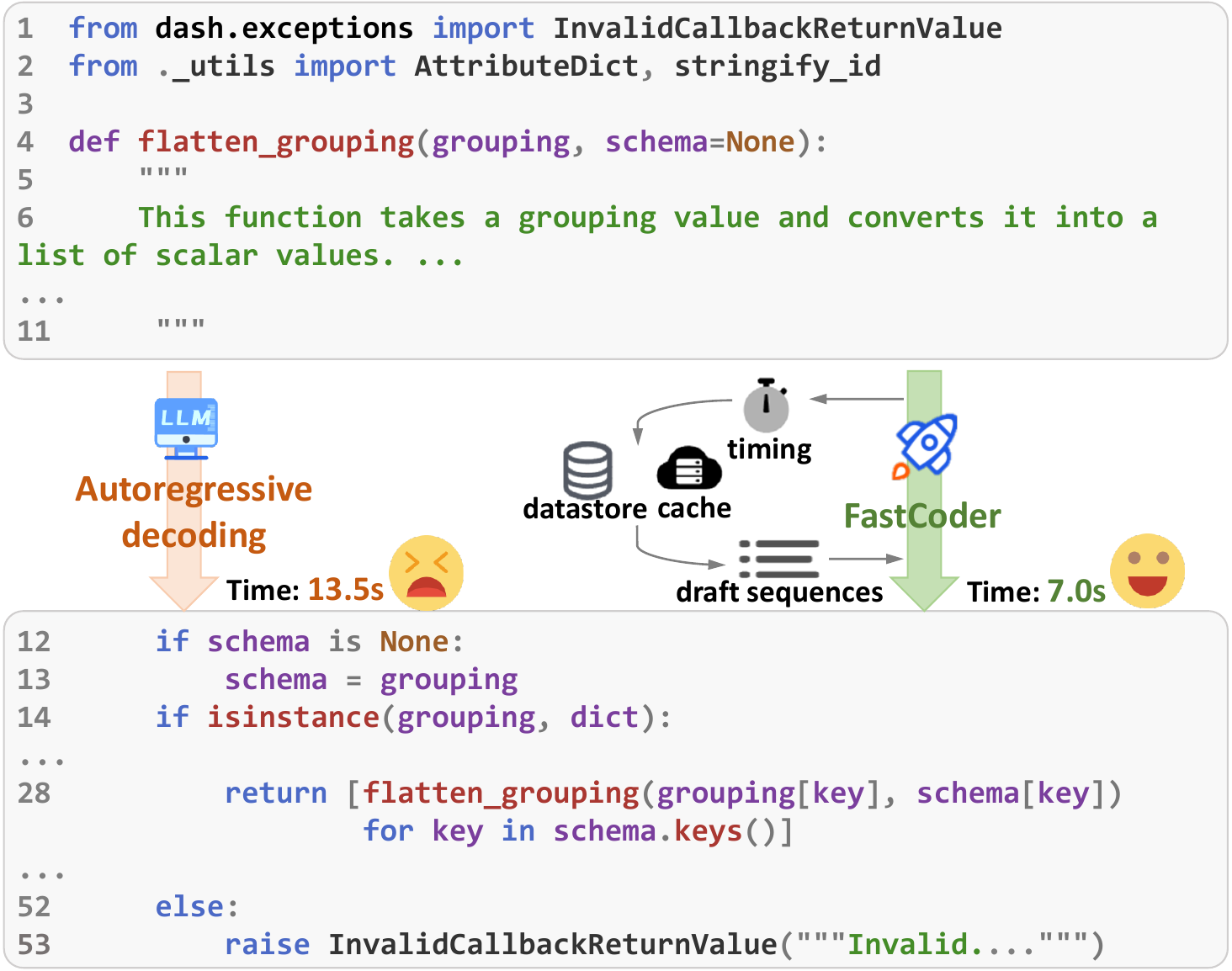}
  \caption{Time overhead of a practical code generation example using CodeLlama-7B on a single NVIDIA 4090 GPU.}
  \label{fig: motivation}
  \vspace{-0.3cm}
\end{figure}

\begin{figure}[t]
 \setlength{\abovecaptionskip}{0.1cm}
    \centering
    \includegraphics[width=0.9\linewidth]{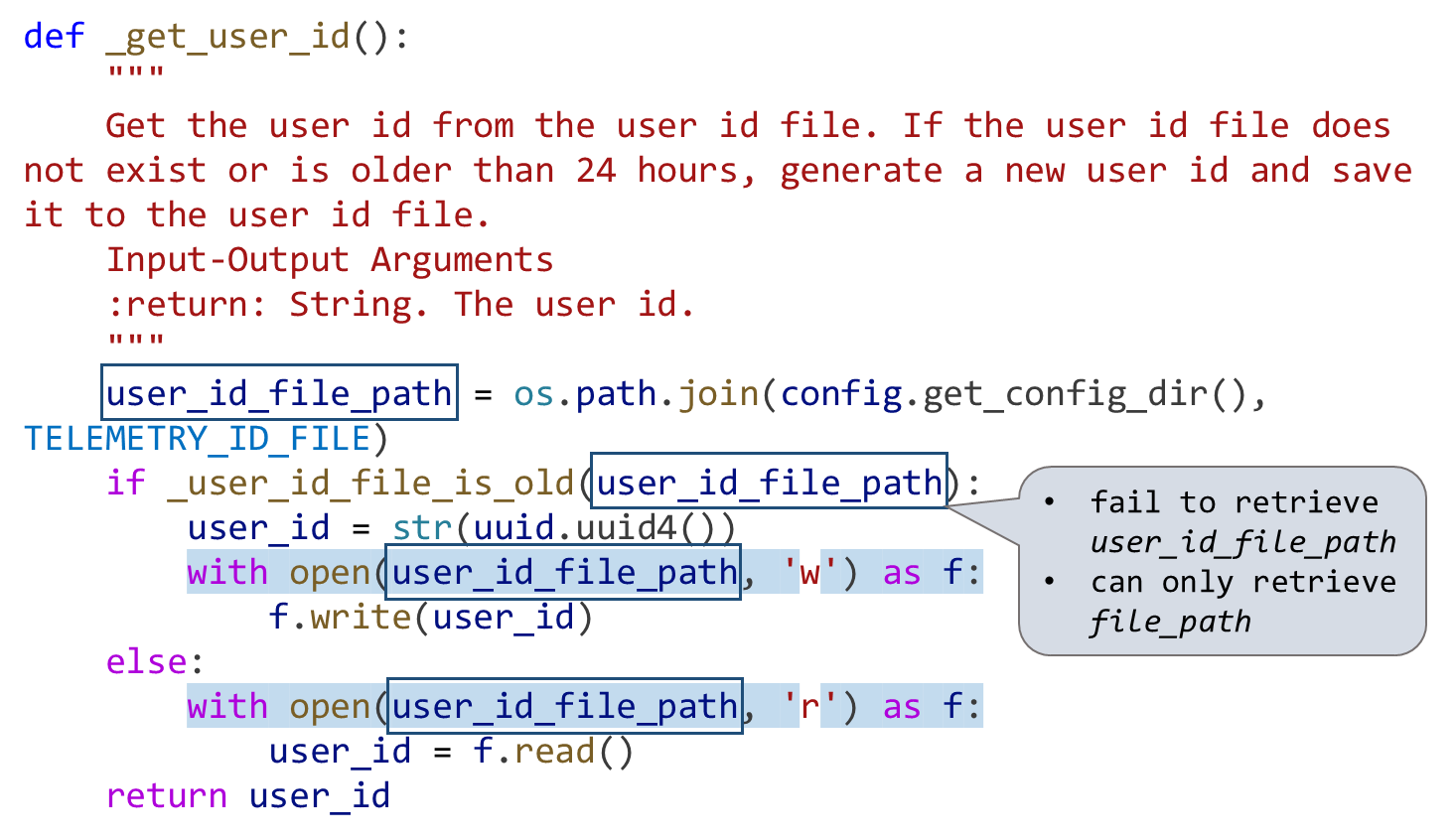}
  \caption{Localness of source code.}
  \label{fig: motivation example}
  \vspace{-0.5cm}
\end{figure}

\subsection{Motivating Examples} \label{sec: observations}
We first present a practical example in Figure \ref{fig: motivation}. Given a function signature and its context as input, autoregressive decoding with an LLM (CodeLlama-7B) takes about 13.5 seconds on a single NVIDIA 4090 GPU to generate the function body, where such latency may negatively affect user experience.
This highlights the need for efficient methods to accelerate code generation.
To identify the limitations of current inference acceleration methods, we present motivating examples that highlight the localness of source code and the retrieval performance in retrieval-based approaches.

\textbf{Localness of source code.}
Human-written programs are typically localized \cite{tu2014localness}, with program entities (token sequences) defined or used in the preceding snippets frequently being reused in the subsequent code snippets within the same code file.
As shown in Fig. \ref{fig: motivation example}, \textit{user\_id\_file\_path} is a user-defined variable within the current code segment, which does not exist in the datastore 
but appears multiple times in subsequent code snippets. Additionally, the blue-highlighted statements demonstrate the repetition of token sequences. \textit{By effectively leveraging these frequently occurring token sequences within the file, such as storing them in a cache for subsequent retrieval, the acceptance length for draft validation can be increased, thereby enhancing the inference speed.}

\begin{figure}[t]
 \setlength{\abovecaptionskip}{0.1cm}
     \begin{subfigure}[b]{0.48\linewidth}
        \includegraphics[width=\linewidth]{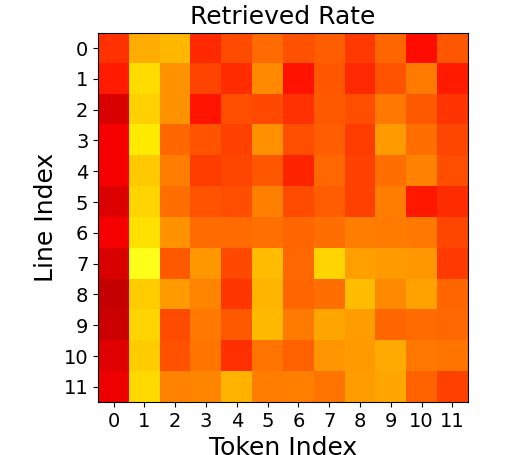}
        \caption{}
    \end{subfigure}
    \begin{subfigure}[b]{0.48\linewidth}
        \includegraphics[width=\linewidth]{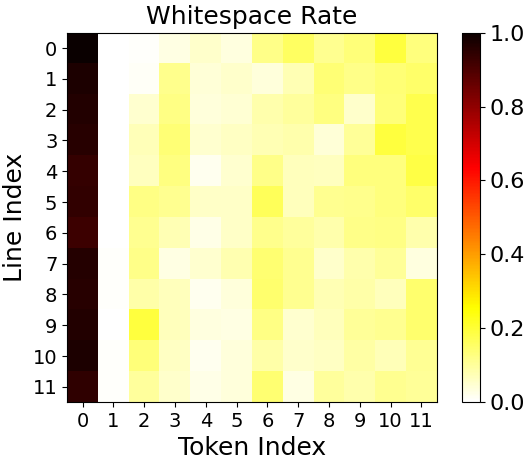}
        \caption{}
    \end{subfigure}
  \caption {Heatmaps of (a) retrieval performance and (b) whitespace distribution with token positions in REST. The maximum token index is selected based on the average token number per line (12).
  }
  \vspace{-0.5cm}
  \label{fig: retrieval analysis}
\end{figure}

\textbf{Retrieval is not always essential.}
Current work performs retrieval operation at \textit{every} position, which may bring unnecessary cost.
To investigate the relationship between retrieval performance and token position in code generation, we randomly selected 200 samples from DevEval \cite{li2024deveval}, a repository-level code generation benchmark, and employed DeepSeek-Coder-6.7B \cite{guo2024deepseek} for evaluation. 
For each token, we recorded whether it was: (a) retrieved from the datastore rather than generated by the model, and (b) a whitespace character (e.g., spaces or newline characters). 
Results are presented as heatmaps in Fig.~\ref{fig: retrieval analysis}. 
As seen from Fig. \ref{fig: retrieval analysis}(a), \textit{retrieval failures are frequent, with a particularly notable pattern: the second token in each line has the lowest probability of being successfully retrieved}.
A comparison with the whitespace rate heatmap suggests that this phenomenon may stem from the fact that the second token is typically the first non-whitespace character at the beginning of a line.
The first non-whitespace token in each line dictates the direction of the line, making it more variable and consequently more challenging to retrieve.
Thus, \textit{skipping retrieval or reducing the retrieval probability at such positions may improve performance}. 

Motivated by the above observations, we propose \ourmodel{}, which is specifically tailored for code generation. As shown in Figure \ref{fig: motivation}, \ourmodel{} significantly reduces the function body generation time from 13.5s to 7.0s.

\begin{figure*}[t]
    \centering
    \setlength{\abovecaptionskip}{0.1cm}
  \includegraphics[width=0.73\linewidth]{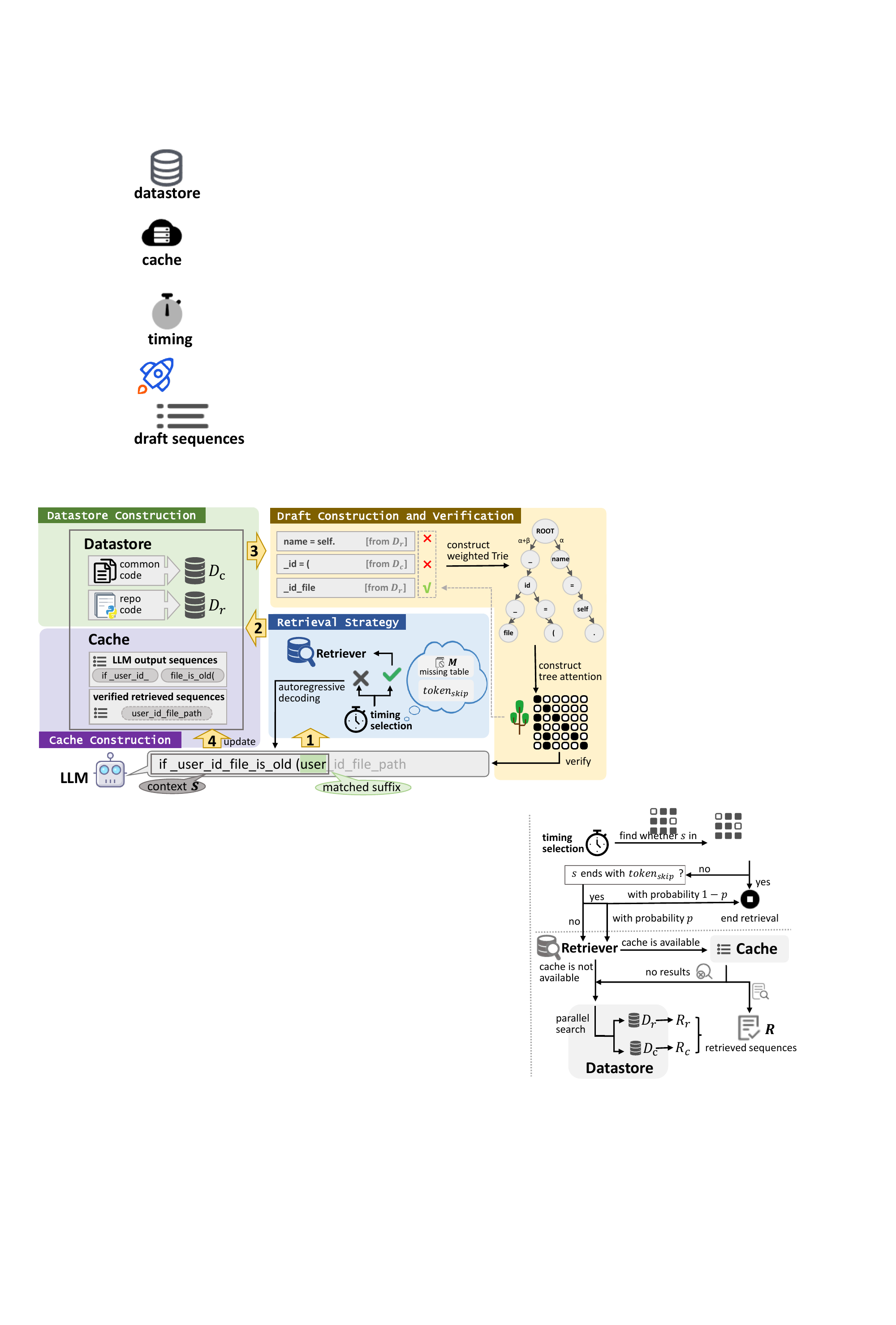}
  \caption{Architecture of \ourmodel{}. 
  } 
  \label{fig: architecture}
  \vspace{-0.5cm}
\end{figure*}

\section{Approach}

To improve the inference efficiency of LLMs on code generation tasks, we propose \ourmodel{}, an LLM inference acceleration approach tailored to the characteristics of code generation.
The architecture of \ourmodel{} is shown in Fig.~\ref{fig: architecture}. 
\ourmodel{} constructs a multi-source datastore combining general and repository-specific knowledge to improve draft quality. It then reduces retrieval overhead by controlling timing, boosts efficiency via parallel retrieval and a context and LLM preference-aware cache. Finally, it uses tree attention to avoid redundant computation.

\subsection{Multi-source Datastore Construction}
The quality of the retrieval datastore, which serves as the source of draft sequences, critically determines the acceleration potential. 
A larger datastore may enhance the probability of result acceptance, but it also correspondingly increases retrieval time, making the trade-off between the two critically important.
To achieve optimal performance with a compact datastore and facilitate effective retrieval, \ourmodel{} incorporates a smaller repository-related datastore $D_r$ and a larger common code datastore $D_c$ to construct a comprehensive retrieval datastore $D$. This design supports parallel retrieval, providing access to both general and project-specific knowledge. To enable fast retrieval with minimal overhead, we organize the datastore into context-continuation pairs, facilitating a rapid exact-match method for context search.

\textbf{Repository-related datastore $D_r$.} During software development, developers often reference cross-file elements such as classes and methods, making intra-repository files highly relevant to the generated code. Additionally, repository-specific factors, including domain variations and coding conventions, lead to distinct patterns of idiomatic expressions. For instance, web development repositories frequently involve HTTP request-response handling, while data science repositories focus on data processing and modeling tasks. To this end, we collect the code files from current repository (with the portions to be generated excluded to avoid data leakage) and form repository-related datastore $D_r$.

\textbf{Common datastore $D_c$.} To ensure that common programming operations are also retrievable, a subset of data from commonly used pre-trained code datasets \cite{kocetkov2022stack} is used to form $D_{c}$, which serves as another component of datastore $D$.

\textbf{Datastore organization.} For efficient retrieval, the datastore is organized as contexts and the corresponding continuations following \cite{he2024rest}.
Specifically, for each code file utilized in constructing the datastore, the content preceding every position will constitute a context, whereas the content subsequent to that position is the corresponding continuation.
The datastore $D$ of \ourmodel{} can be summarized as:
\setlength{\abovedisplayskip}{3pt}
\setlength{\belowdisplayskip}{3pt}
\begin{align}
     D &= ( D_{r}, D_{c} ) \\
    (D_{r}, D_{c}) &= ( \{(c_i, t_i)\}_{i=1}^{|D_{r}|}, \{(c_j, t_j)\}_{j=1}^{|D_{c}|} )
\end{align}
where $c_i$ ($c_j$) represents the context, $t_i$ ($t_j$) represents the corresponding continuation of $c_i$ ($c_j$), $|D_{r}|$ ($|D_{c}|$) is the number of samples in $D_r$ ($D_c$).
Specifically, $D_r$ can be omitted in standalone code generation where such context is unreachable.

\subsection{Context- and LLM Preference-aware Caching}
To reduce retrieval costs and improve the alignment of retrieved results with context and LLM preferences—thereby increasing both the accepted sequence length and inference speed—we design a context- and LLM preference-aware caching strategy to cache the verified retrieved sequences and LLM-generated sequences.

Based on the observations in Section \ref{sec: observations} that program entities (token sequences) defined or used in preceding snippets are often reused in the subsequent code snippets, we design the \textsc{cache} mechanism from two perspectives.
On the one hand, although datastore $D$ contains vast content, typically only a small portion is highly relevant to the code currently being generated.
In contrast, the validated retrieval tokens are exactly the sequences that exhibit high relevance to the current generation code, and are thus more likely to be retrieved again in subsequent stages than other content within the datastore $D$.
Consequently, if the draft sequence $r=(y_1,...,y_j)$, retrieved by the context $s=(x_1,...,x_t)$, is verified by the LLM, we concatenate them as $(x_1,...,x_t,y_1,...,y_j)$ and add it into \textsc{cache}.
On the other hand, since the datastore $D$ is static, as it remains unmodified after construction, the draft sequences retrieved for the identical context $s$ also remain consistent.
However, different LLMs exhibit distinct generation preferences, which is reflected in the fact that they may generate different outputs given the same input.
As a result, a static datastore struggles to accommodate the diverse preferences of various LLMs.
Earlier decoding outputs can to some extent reflect LLM-specific tendencies and ensure contextual coherence. Therefore, we also incorporate the verified decoding output sequence into \textsc{cache} for future use.

To maintain the \textsc{cache}, we assess whether the two aforementioned update conditions are satisfied after each forward step of the LLM. If the number of sequences inside the \textsc{cache} exceeds the pre-defined threshold $l$, it is accessible and will remain active throughout the entire inference process.

\subsection{Dynamic and Efficient Retrieval Strategy}

Algorithm \ref{retrieval algorithm} illustrates the complete retrieval process of \ourmodel{}.
Before each forward step, given current context $s$, \ourmodel{} initially verifies the availability of \textsc{cache}. If the \textsc{cache} is accessible, that is, the number of sequences inside exceeds $l$, retrieval is prioritized from \textsc{cache}. 
If \textsc{cache} is unavailable or fails to yield valid (non-empty) results, \ourmodel{} utilizes a dynamic and efficient retrieval strategy to minimize unnecessary retrieval cost. 
Specifically, \ourmodel{} optimizes retrieval timing by addressing two key considerations as follows.

\textbf{Skip token.}
As mentioned in Section \ref{sec: observations}, the intrinsic characteristics of code lead to a low retrieval success rate at the first non-whitespace character of each line. 
Since obvious patterns are not found in other positions, and the introduction of intricate judgment processes may incur additional computational overhead, we set the first non-whitespace character of each line as the skip token.
We strategically reduce the retrieval probability of skip tokens through a control parameter $p$, which refers to the retrieval probability at these positions.

\textbf{Missing table.}
When utilizing the current context $s$ to retrieve its continuations from datastore $D$, it may fail to yield any valid results in some cases. 
To prevent time wastage resulting from invalid retrieval, we maintain a missing table $M=\{s_{m_i}\}$ that stores suffixes $s_{m_i}$ for which no valid results can be retrieved from the datastore $D$.
Thus, when $s_{m_i}$ is encountered again during the subsequent inference, \ourmodel{} will bypass the retrieval and directly utilize the LLM to generate the next token. 

If \ourmodel{} decides to proceed with retrieval according to the above strategy, parallel retrieval is conducted from $D_{r}$ and $D_{c}$ to further boost the retrieval efficiency, and the results refer to $R_{r}$ and $R_{c}$, separately. Specifically, if $R_{r}$ and $R_{c}$ are both empty, $s$ will be denoted as $s_m$ and added into the missing table $M$. Otherwise, relevant sequences are employed to update the \textsc{cache}.

\setlength{\textfloatsep}{6pt}
\begin{algorithm}[t]
\small
\caption{Retrieval Algorithm}
\label{retrieval algorithm}
\KwIn{current context $s$, datastore $D$, retrieval cache \textsc{cache}, minimum activation size $l$, missing table $M$, skip token $\textit{token}_\text{skip}$, retrieval probability $p$}
\KwOut{Retrieved sequences $R$}

\If{$\textsc{cache}.size \geq l$}{ 
    // retrieval from cache\\
    $R \gets \text{search(\textsc{cache})}$
}
\If{$\textsc{cache}.size < l$ \textbf{or} $R=\emptyset$}{
    // retrieval timing selection \\
    \If{$s \in M$}{
        pass
    }
    \ElseIf{$s$ ends with $\textit{token}_\text{skip}$}{
        \If{random number  $< p$}{
            // parallel retrieval from datastore\\
            $R_{r}, R_{c} \gets \text{par\_search}(D_{r}, D_{c})$ \\
            $R \gets (R_{r}, R_{c})$
        }
    }
}
\If{$R = \emptyset$}{
    // update missing table \\
    $M \gets M \cup \{s\}$
}
\Else{
    update \textsc{cache}
}
\Return $R$\;
\end{algorithm}

\subsection{Draft Construction and Verification with Weighted Prefix Optimization} 
The retrieval results $R =(R_{r}, R_{c})$ contain potential continuations of the current context $s$, often sharing the same prefix. To reduce the cost brought by verification each $r_i \in R$ one by one, we construct the draft sequences using a Trie, where the unique path from a node to the root node corresponds to a prefix of the retrieval results, aiming to reduce the repeated verification of shared prefixes in $R$. 
We use following equation to assign a weight for each node:
\setlength{\abovedisplayskip}{3pt}
\setlength{\belowdisplayskip}{3pt} 
\begin{align}
    N_{weight} = \alpha \cdot t_{r} + \beta \cdot t_{c}
\end{align}
where $t_{r}$ and $t_{c}$ represents the times that the node occurs in $R_{r}$ and $R_{c}$ respectively, and  $\alpha$ and $\beta$ refers to the corresponding coefficient.
We retain $\alpha$ and $\beta$ as tunable parameters to accommodate diverse scenarios.
For instance, in highly specialized code generation tasks, increasing the $\alpha/\beta$ value can help generate draft sequences that better align with repository-specific characteristics. 
We select top-$k$ weighted sequences from the Trie as the draft sequences.

As many draft sequences may share common prefixes, to avoid redundant computation of Transformer layers, we employ tree attention \cite{spector2023treeattention, miao2024specinfer} to verify the draft sequences.
We represent the tree formed by the draft sequences, which is part of the Trie, as $\mathcal{T}=(V, E)$, where $V$ is the set of all generated tokens and $E$ is the set of edges representing token transitions in the candidates. The total number of nodes is $ N=|V|$.  
In tree attention, only a token's predecessors are considered as historical context, and the attention mask restricts attention to these predecessors. 
Let $\mathbf{A} \in \{0,1\}^{N \times N}$ be the attention mask matrix, where $A_{ij}=1$ indicates that token $i$ can attend to token $j$. To reflect the tree structure, we define:
\begin{equation}
A_{ij} =
\begin{cases}
1, & \text{if token } j \text{ is a predecessor of token } i \text{ in } \mathcal{T} \\
0, & \text{otherwise}
\end{cases}    
\end{equation}
This ensures that each token can only attend to its own continuation path, and tokens from different candidate branches are isolated in the attention computation.

By applying tree attention mask and appropriately adjusting the positional indices for encoding, we are able to process multiple candidates simultaneously without increasing the batch size.
As our objective is to accelerate the inference without compromising model performance, all correct tokens from the beginning will be accepted, while the draft tokens following the first error will be rejected.

\section{Evaluation}

To evaluate the effectiveness of \ourmodel{}, we address the following research questions:

\vspace{-0.15cm}
\begin{center}\small
\begin{tcolorbox}[colback=gray!10,
                  colframe=black,
                  arc=1mm, auto outer arc,
                  boxrule=0.5pt, breakable
                 ]
\textbf{RQ1: Overall Performance.}
How does the inference acceleration performance of \ourmodel{} compare against state-of-the-art approaches? 

\textbf{RQ2: Ablation Study.}
What are the contributions of each component of \ourmodel{}?

\textbf{RQ3: Integration with Correctness-Focused Approaches.}
Can \ourmodel{} enhance generation speed when integrated with existing code generation methods that focus on the code correctness improvement?

\end{tcolorbox}
\end{center}

\subsection{Datasets and Backbone Models.}
\ourmodel{} is not only applicable to repository-level code generation, but can also be adapted to standalone code generation with minimal modifications.
Thus, we conduct experiments on both repository-level and standalone code generation benchmarks. For repository-level code generation, we choose two widely-used benchmarks, DevEval \cite{li2024deveval} and RepoEval \cite{zhang2023repocoder}.
DevEval comprises 1,825 testing samples from 115 repositories, covering 10 popular domains. It aligns with real-world repositories in code distributions and dependency distributions.
RepoEval is constructed using the high-quality repositories sourced from GitHub.
We use the function-level subset for evaluation, which contains 455 testing samples.
For standalone code generation, we conduct experiments on HumanEval \cite{chen2021codex}, a widely-used standalone code generation dataset including 164 human-written programming problems.

For backbone models, we use the 1.3B and 6.7B configurations of Deepseek-Coder-base \cite{guo2024deepseek}, as well as 7B and 13B configurations of CodeLlama-Python \cite{roziere2023codellama} for evaluation, which are popular and well-performing LLMs in code generation.

\subsection{Baselines.}
We compare \ourmodel{} with vanilla \textbf{Autoregressive decoding} and the following state-of-the-art inference acceleration approaches that follow the draft-verification framework and have demonstrated effectiveness in code generation:
\begin{itemize}[leftmargin=*]
    \item \textbf{Self-speculative decoding }\cite{zhang2024draft}: This approach is generation-based and employs the target LLM with selectively certain intermediate layers skipped as the draft model, without the need for an auxiliary model.
    \item \textbf{Ouroboros} \cite{zhao2024ouroboros}: This is a generation-based approach and demands manual selection of a suitable draft model for the target LLM. To maximize the utility of discarded drafts, it generates phrases from them to parallelize the drafting process and lengthen drafts in a training-free manner.
    \item \textbf{REST} \cite{he2024rest}: REST is a retrieval-based and draft model-free approach, which draws from the reservoir of existing knowledge, retrieving and employing relevant tokens based on the current context.
\end{itemize}

\subsection{Evaluation Metrics.}
Following existing work \cite{zhang2024draft, zhao2024ouroboros, he2024rest}, we adopt the following metrics to assess the inference efficiency:
\begin{itemize}[leftmargin=*]
    \item \textbf{Decoding speed (ms/token)}: The average generation time of one token for the LLM with batch size set to 1 on a single GPU. Formally, if $L$ denotes the length of the generated tokens and $T$ represents the time spent for generation, decoding speed is computed by:
    \begin{equation}
        \textit{Decoding Speed} = T/L
    \end{equation}
    \item \textbf{Speedup}: The ratio by which the decoding speed of an evaluated method is increased compared to vanilla autoregressive decoding. Specifically, if $Speed_{A}$ and $Speed_{B}$ represent the decoding speed of vanilla autoregressive decoding and the evaluated approach, respectively, the speedup can be calculated by:
    \begin{equation}
        \textit{Speedup} = Speed_A / Speed_B
    \end{equation}
    \item \textbf{Average Acceptance Length}: The average number of tokens accepted per forward step by the target LLM, which represents the theoretical upper bound of achievable acceleration for retrieval-based approaches. Formally, if $L$ denotes the length of generated tokens, and $F$ represents the number of forward steps by the target LLM, it can be calculated by:
    \begin{equation}
        \textit{Average Acceptance Length} = L/F
    \end{equation}
\end{itemize}

Although \ourmodel{} can generate code theoretically consistent with the target LLM's output, we still evaluate the correctness of the generated code to enable quantitative assessment.
For DevEval \cite{li2024deveval} and HumanEval \cite{chen2021codex}, we report \textbf{Pass@1} results.
For RepoEval \cite{zhang2023repocoder}, we report \textbf{Edit Similarity (ES)} results, since Pass@k scripts are unavailable due to licensing as stated in their GitHub repository.

\begin{table*}[t]
    \centering
    \setlength{\abovecaptionskip}{0.1cm}
    \caption{Decoding speed and speedup ratio on repository-level code generation datasets. For \ourmodel{}, we additionally report the \textit{average per-sample decoding time} and \textit{Pass@1 / ES score}, with the comparison to autoregressive decoding shown in the upper-right corner, which are highlighted with a blue background (\comparison{$Avg. Time / Pass@1 ^{\Delta v.s.AR}$} for DevEval and \comparison{$Avg. Time / ES ^{\Delta v.s.AR}$} for RepoEval).}
    \resizebox{\linewidth}{!}{ 
    \begin{tabular}{l|l|cc|cc|cc|cc}
        \toprule
        \multirow{2}{*}{\textbf{Dataset}} & \multirow{2}{*}{\textbf{Approach}} & \multicolumn{2}{c|}{\textbf{Deepseek-Coder-1.3B}} & \multicolumn{2}{c|}{\textbf{Deepseek-Coder-6.7B}} & \multicolumn{2}{c|}{\textbf{CodeLlama-7B}} & \multicolumn{2}{c}{\textbf{CodeLlama-13B}} \\
         & &  \textbf{ms/token} & \textbf{Speedup} &  \textbf{ms/token} & \textbf{Speedup} &  \textbf{ms/token} & \textbf{Speedup} &  \textbf{ms/token} & \textbf{Speedup}\\
        \midrule
        \multirow{6}{*}{DevEval} & Autoregressive & 20.00 & 1.00$\times$ & 26.15 & 1.00$\times$ & 26.29 & 1.00$\times$ & 46.35 & 1.00$\times$\\
         & Self-speculative & 18.72 & 1.07$\times$ & 22.55 & 1.16$\times$ & 25.10 & 1.05$\times$ & 42.74 & 1.08$\times$ \\
         & Ouroboros & - & - & 15.69 & 1.67$\times$ & 29.14 & 0.90$\times$ & \underline{39.73} & \underline{1.17$\times$} \\
         & REST & \underline{12.10} & \underline{1.65$\times$} & \underline{15.28} & \underline{1.71$\times$} & \underline{15.57} & \underline{1.69$\times$} & 43.38 & 1.07$\times$ \\
         & \ourmodel{} & \textbf{8.71} & \textbf{2.30$\times$} & \textbf{11.69} & \textbf{2.24$\times$} & \textbf{12.17} & \textbf{2.16$\times$} & \textbf{21.56} & \textbf{2.15$\times$}\\
        & & \multicolumn{2}{c|}{\comparison{$4.4s^{\downarrow5.8s}$ / $20.38^-$}} & \multicolumn{2}{c|}{\comparison{$6.0s^{\downarrow7.4s}$ / $27.01^-$}} & \multicolumn{2}{c|}{\comparison{$6.2s^{\downarrow7.3s}$ / $29.81^-$}} & \multicolumn{2}{c}{\comparison{$11.0s^{\downarrow12.7s}$ / $30.90^-$}}\\

        \midrule
        \multirow{6}{*}{RepoEval} & Autoregressive & 19.91 & 1.00$\times$ & 25.75 & 1.00$\times$ & 26.21 & 1.00$\times$ & 47.86 & 1.00$\times$\\
         & Self-speculative & 19.63 & 1.02$\times$ & 22.48 & 1.16$\times$ & 24.36 & 1.08$\times$ & 42.09 & 1.14$\times$ \\
         & Ouroboros & - & - & \underline{14.56} & \underline{1.77$\times$} & 33.12 & 0.79$\times$ & \underline{35.60} & \underline{1.34$\times$} \\
         & REST & \underline{12.09} & \underline{1.65$\times$} & 15.46 & 1.67$\times$ & \underline{15.43} & \underline{1.70$\times$} & 44.59 & 1.04$\times$ \\
         & \ourmodel{} & \textbf{7.88} & \textbf{2.53$\times$} & \textbf{10.83} & \textbf{2.38$\times$} & \textbf{10.80} & \textbf{2.43$\times$} & \textbf{19.02} & \textbf{2.52$\times$}\\
         & & \multicolumn{2}{c|}{\comparison{$4.0s^{\downarrow6.2s}$ / $34.33^-$}} & \multicolumn{2}{c|}{\comparison{$5.5s^{\downarrow7.7s}$ / $37.79^-$}} & \multicolumn{2}{c|}{\comparison{$5.5s^{\downarrow7.9s}$ / $38.17^-$}} & \multicolumn{2}{c}{\comparison{$9.7s^{\downarrow14.8s}$ / $37.58^-$}}\\
        \bottomrule
    \end{tabular}
    }
    \label{tab: main results}
    \vspace{-0.3cm}
\end{table*}

\begin{figure*}[t]
 \centering
  \setlength{\abovecaptionskip}{0.1cm}
  \includegraphics[width=\linewidth]{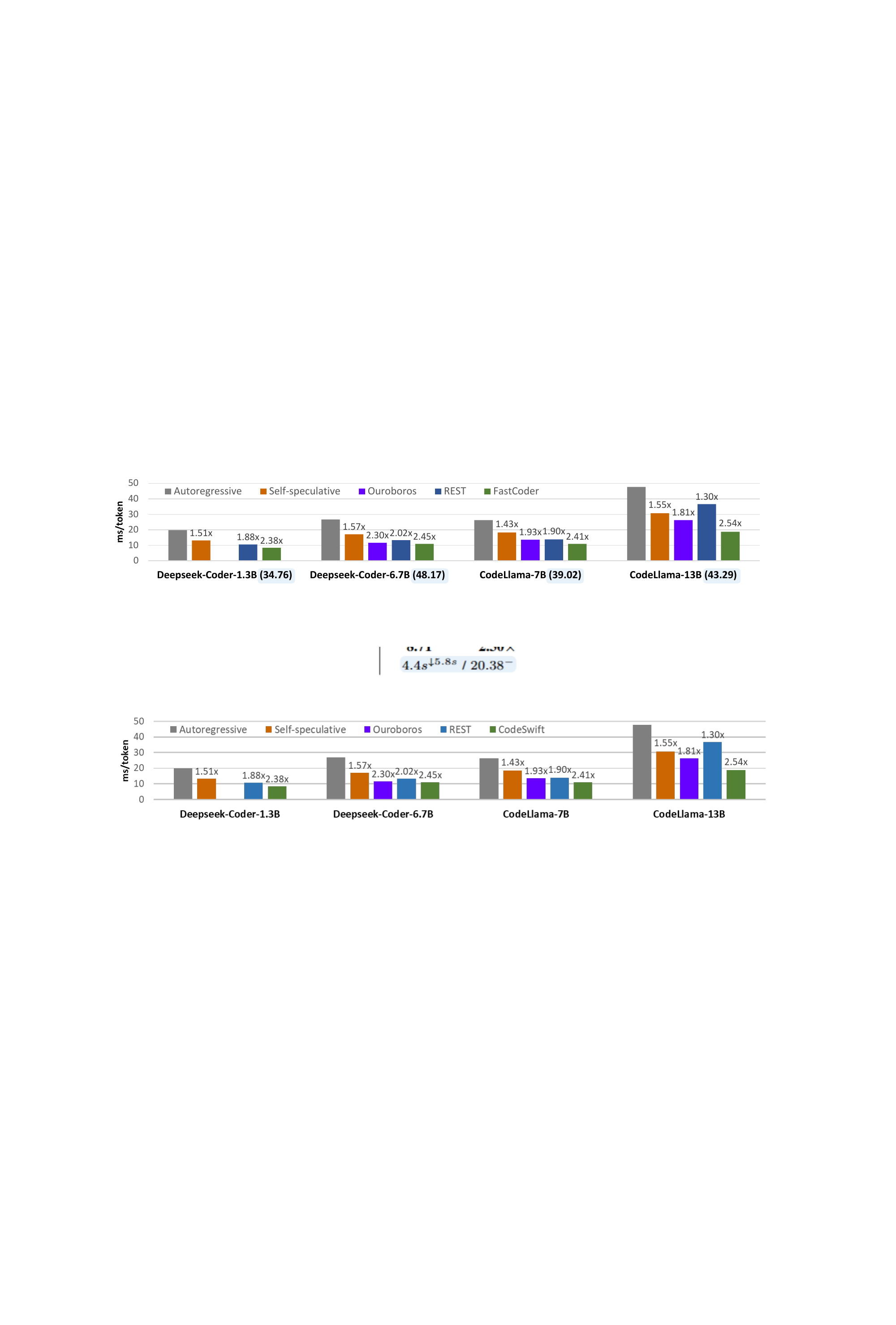}
  \caption{Decoding speed and speedup ratio on standalone code generation dataset HumanEval. We also report the \textit{Pass@1} results of \ourmodel{} in parentheses with a blue background, which are the same with autoregressive decoding.}
  \label{fig: humaneval results}
  \vspace{-0.5cm}
\end{figure*}

\subsection{Implementation Details.} \label{sec:impl}
To provide essential contextual information, we prepend preceding code snippets from the same file as context for DevEval and RepoEval.
Following existing works \cite{zhang2023repocoder, li2024deveval, he2024rest}, all results are obtained with a maximum input length of 2k and generation length of 512 under greedy decoding.
$D_{c}$ is constructed from the first Python file of pre-training code in The Stack \cite{kocetkov2022stack}, taking approximately 9 minutes and yielding a 0.9GB datastore.
$D_{r}$ ranges from 60KB to 289MB across repositories, taking an average of 10 seconds.
Based on preliminary experiments, we set the hyper-parameters to $l=50$, $p=0.5$, and $\alpha=\beta=1$, with LLM output truncated every 20 tokens and added to the \textsc{cache}. Evaluation of $p$ over the range 0.1 to 0.9 and $l$ from 10 to 100 showed an initial improvement followed by a decline, and the final values were chosen for best performance. $\alpha$ and $\beta$ were similarly selected from a range of tested combinations.
Following \cite{he2024rest}, for retrieval, the starting context suffix length $n_{max}=16$, and a maximum of 64 draft tokens of the top-$k$ sequences are selected in the Trie.
All experiments with Deepseek-Coder-1.3B/6.7B and CodeLlama-7B use a single NVIDIA 4090 GPU and 28 CPU cores, and CodeLlama-13B experiments use a single NVIDIA A6000 GPU and 12 CPU cores.

The implementation details of baseline approaches can be seen as follows.
For Self-speculative decoding, we adopt the skipped layers from \cite{zhang2024draft} for CodeLlama-13B. For DeepSeek-Coder and CodeLlama-7B, where specific layers are not provided, we follow the instruction to determine them, which takes several hours. 
For Ouroboros, while selecting the draft model, we prioritize the smaller model from the same series as the target LLM. Thus, we choose Deepseek-Coder-base-1.3B for Deepseek-Coder-base-6.7B, and CodeLlama-Python-7B for CodeLlama-Python-13B as draft models. For CodeLlama-7B, which is the smallest model in its series, we opt for TinyLlama-1.1B as the draft model due to its shared architecture and tokenizer compatibility.
Due to the small size of DeepSeek-Coder-1.3B, identifying a suitable draft model for it is challenging, and therefore, results for this setting are not reported.
For REST, to construct the datastore, we select the first 10 files in Python subset of The Stack dataset \cite{kocetkov2022stack}, resulting in a datastore of approximately 8.7 GB in size.
The values of the other hyper-parameters are consistent with those in the original paper.

\section{Results and Analysis}
\subsection{RQ1: Overall Performance}

To answer this research question, we compare the inference acceleration performance of \ourmodel{} against state-of-the-art approaches on both repository-level and standalone code generation tasks, and the detailed results and analysis are presented as follows.

\subsubsection{Repository-level Code Generation}
The comparison results between \ourmodel{} and baselines are shown in Table~\ref{tab: main results}.
\ourmodel{} achieves up to $2.30 \times$ and $2.53 \times$ speedup on DevEval and RepoEval, respectively, 
outperforming state-of-the-art approaches by up to $88\%$.
\ourmodel{} consistently maintains a stable speedup of more than $2 \times$ across a variety of backbone models and datasets, demonstrating its robustness.
On average, \ourmodel{} saves 12.7s per sample on DevEval and 14.8s per sample on RepoEval compared with autoregressive decoding.
The Pass@1 and Edit Similarity scores further confirm that \ourmodel{} produces sequences identical to those of autoregressive generation, as \ourmodel{} accepts draft tokens only when they match the target LLM’s outputs.

Compared to the substantial speedups gained by \ourmodel{}, baseline approaches achieve limited accelerations.
As a retrieval-based approach, the datastore utilized by REST is approximately 8 times the size of the one employed by \ourmodel{}, but the fixed datastore may suffer from misalignment between retrieval sequences and the LLM output.
REST exhibits the optimal speedup of around $1.7\times$ in most cases, but it performs poorly in experiments of CodeLlama-13B. 
This may be attributed to the fact that the significant CPU resource demands posed by both the 13B model inference and the retrieval of data from a large datastore in REST, leading to decreased performance.
Besides, Ouroboros demonstrates comparable performance to REST on Deepseek-Coder-6.7B, yet its generation speed is even slower than autoregressive decoding on CodeLlama-7B, indicating that its efficacy is subject to considerable fluctuations influenced by factors such as model selection, further demonstrating the challenge of appropriate draft model selection.
Self-speculative decoding consistently maintains a stable yet modest acceleration, and the preliminary search for skipped layers may introduce significant time overhead.
In contrast, \textbf{\ourmodel{} does not require a draft model or additional training, yet it can maintain a stable speedup ratio even under resource-constrained conditions.}

\subsubsection{Standalone Code Generation}
For \ourmodel{}, we remove $D_r$ from the datastore and retain $D_c$, which is the same as the one used in the previous experiments. 
The results are shown in Fig. \ref{fig: humaneval results}. \textbf{Even without the benefit of the multi-source datastore, \ourmodel{} still outperforms the baselines}, further demonstrating the effectiveness of the retrieval strategy and caching modules.
The identical Pass@1 scores between \ourmodel{} and autoregressive decoding indicate that \ourmodel{} achieves acceleration without compromising model performance.
Additionally, we observe that the baselines consistently perform better on HumanEval compared to repository-level datasets.
This may be affected by the difference in difficulty between standalone and repository-level code generation tasks.
For instance, Deepseek-Coder-1.3B achieves pass@1 scores of 34.8 on HumanEval and 18.2 on DevEval.
Thus, for approaches such as Ouroboros and Self-speculative, which require a draft model, the performance in repository-level code generation may be negatively affected by the poor performance of the draft model.
For REST, HumanEval involves no project-specific knowledge, and the common datastore may adequately satisfy retrieval requirements.
The performance differences of existing approaches on the two types of code generation tasks also highlight that \textit{evaluations based solely on standalone datasets may fail to reflect performance in real-world application scenarios.}

\vspace{1mm}
\begin{custommdframed}
\textbf{Answer to RQ1:} \ourmodel{} achieves state-of-the-art performance on both repository-level and standalone code generation tasks. It consistently maintains a stable speedup across a variety of backbone models and datasets, demonstrating its effectiveness and generalization.
\end{custommdframed}
\vspace{-0.3cm}

\subsection{RQ2: Ablation Study} \label{sec:ablation}

To analyze the effectiveness of each component within \ourmodel{}, we conduct an ablation study.
Since REST serves as a standard pipeline for retrieval-based draft-verification acceleration approach, we treat the results obtained using REST (with $D_c$ as the datastore) as the baseline results.
To ensure a thorough analysis, we first evaluate the performance of the baseline with each individual component added separately, followed by an evaluation of the baseline with any two components combined.

Table \ref{tab: ablation} reports the results for \textit{decoding speed} and \textit{average acceptance length}, demonstrating that each component contributes to a speedup gain.
The multi-source datastore improves retrieval performance by offering richer and more interrelated content. 
This performance also represents REST using the same datastore ($D_r+D_c$) as \ourmodel{}, along with parallel retrieval.
It indicates that the multi-source datastore design not only increases the average acceptance length but also reduces the external retrieval overhead through efficient parallel search mechanisms.
The retrieval strategy optimizes the inference speed by eliminating unnecessary retrieval operations, which account for approximately 4\% of the total retrievals. Crucially, this optimization has a negligible impact on the average acceptance length, ensuring minimal trade-off between efficiency and performance.
Among all components, the \textsc{cache} mechanism proves to be the most effective, delivering a substantial 30\%+ improvement in average acceptance length over the baseline.
Statistical analysis shows that, although the \textsc{cache} contains only 174 sequences at most for DevEval, 33.13\% of all retrieval operations can successfully obtain valid results directly from the \textsc{cache}. 
Besides, the average retrieval time from the cache is 0.2ms, which is approximately 15\% of the retrieval time from the datastore $D$.
This highlights the importance of alignment with the repository context and LLM preference.
Furthermore, the combination of any two components consistently outperforms their individual effects, while integrating all three components achieves optimal performance, confirming positive interactions between components.

\begin{table}[t]
\setlength{\abovecaptionskip}
{0.1cm}
    \caption{Ablation study results of \ourmodel{} on DevEval using Deepseek-Coder-6.7B. 
    \textit{AccLen} refers to average acceptance length.}
    \label{tab: ablation}
    \centering
    \resizebox{\linewidth}{!}{ 
    \begin{tabular}{lccc}
        \toprule
         & \textbf{AccLen} & \textbf{ms/token} & \textbf{Speedup} \\
        \midrule
        Baseline & 1.89 & 15.86 & 1.65$\times$\\
        \midrule
        + datastore $D$ ($D_r+D_c$) & 2.28 & 14.82 & 1.76$\times$\\
        + retrieval strategy & 1.89 & 15.41 & 1.70$\times$\\
        + \textsc{cache} & 2.47 & 14.13 & 1.85$\times$\\
        \midrule
        + datastore $D$ \& retrieval strategy & 2.28 & 14.19 & 1.84$\times$\\
        + retrieval strategy \& \textsc{cache} & 2.48 & 13.80 & 1.89$\times$\\
        + datastore $D$ \& \textsc{cache} & 2.85 & 11.94 & 2.19$\times$\\
        \midrule
        \textbf{\ourmodel{}} & \textbf{2.85} & \textbf{11.69} & \textbf{2.24$\times$}\\
        \bottomrule
    \end{tabular}
    }
    \label{tab: ablation}
\end{table}

\vspace{1mm}
\begin{custommdframed}
\textbf{Answer to RQ2:} Each component of \ourmodel{} makes a contribution to the overall performance, with the \textsc{cache} mechanism proving to be the most effective, providing a substantial improvement in average acceptance length and speedup ratio. 
\end{custommdframed}
\vspace{-0.3cm}

\subsection{RQ3: Integration with Correctness-Focused Approaches}

To evaluate whether \ourmodel{} can benefit existing correctness-focused code generation approaches, we select RepoCoder \cite{zhang2023repocoder} and RLCoder \cite{wang2024rlcoder} for experiments, both of which follow RAG strategy.
These approaches can be summarized as a two-phase process: \ding{172} retrieval of relevant code snippets from the repository through an optimized retrieval strategy to facilitate informative \textit{prompt construction}, followed by \ding{173} \textit{code generation} via LLM inference using the constructed prompt.
Since \ourmodel{} solely accelerates LLM generation while preserving output consistency, it can be seamlessly integrated into the code generation phase without modifying the prompt construction stage from the original approach.

The speedup results can be seen in Table \ref{tab: integrated performance}, including the time spent on each phrase as well as the total execution time (seconds). 
As RepoCoder utilizes an iterative retrieval-generation pipeline, we adopt the optimal configuration (iterations = 2) from the original work and report the cumulative time consumption for each phase.
From the results, we can observe that, with prompt construction time held constant, \ourmodel{} substantially reduces the time required during the generation phase, achieving speedups of $2.68 \times$ and $2.66 \times$ on RepoCoder and RLCoder, respectively.
To quantitatively assess code correctness, we also report Edit Similarity results in Table \ref{tab: integrated performance}, which demonstrates that \ourmodel{} and autoregressive decoding show consistent performance.
This indicates high compatibility of \ourmodel{}, achieving inference speed acceleration through flexible integration with diverse existing code generation approaches following the RAG strategy while maintaining their correctness performance. 

\begin{table}[t]
    \caption{\textit{Speedup} and \textit{ES} results of integration of \ourmodel{} with existing correctness-focused code generation approaches on RepoEval using Deepseek-Coder-1.3B.}
    \label{tab: integrated performance}
    \centering
    \resizebox{\linewidth}{!}{ 
    \begin{tabular}{l|ccc|c|c}
        \toprule
         \multirow{2}{*}{\textbf{Approach}} & \multicolumn{3}{c|}{\textbf{Time (s)}} & \multirow{2}{*}{\textbf{Speedup}} & \multirow{2}{*}{\textbf{ES}}\\
         & \textbf{Prompt}  & \textbf{Generation} & \textbf{Total} & \\
        \midrule
        RepoCoder & \multirow{2}{*}{91.17} & 11,731.88 & 11,823.05 & 1.00$\times$ & 30.35\\
        \textbf{+ \ourmodel{}} &  & 4,326.71 & 4,408.88 & \textbf{2.68$\times$} & 30.35\\ 
        \midrule
        RLCoder & \multirow{2}{*}{204.11} & 5,419.50 & 5,623.61 & 1.00$\times$ & 38.28\\
        \textbf{+ \ourmodel{}} &  & 1,908.78 & 2,112.89 & \textbf{2.66$\times$} & 38.28\\
        \bottomrule
    \end{tabular}
    }
\end{table}

\begin{table}[t]
    \caption{Comparison of average acceptance length between \ourmodel{} and REST. \textit{DE}, \textit{RE}, and \textit{HE} denote DevEval, RepoEval, and HumanEval respectively.}
    \label{tab: acclen}
    \centering
    \resizebox{\linewidth}{!}{
    \begin{tabular}{l|ccc|ccc}
        \toprule
        \multirow{2}{*}{\textbf{Bckbone Model}} & \multicolumn{3}{c|}{\textbf{REST}} & \multicolumn{3}{c}{\textbf{\ourmodel{}}} \\ 
        & \textbf{DE} & \textbf{RE} & \textbf{HE} & \textbf{DE} & \textbf{RE} & \textbf{HE} \\ 
        \midrule
        Deepseek-Coder-1.3B & 2.04 & 2.04 & 2.38 & 2.97 & 3.21 & 2.87 \\
        Deepseek-Coder-6.7B & 2.06 & 2.08 & 2.38 & 2.85 & 3.05 & 2.92 \\
        CodeLlama-7B & 2.05 & 2.07 & 2.27 & 2.77 & 3.06 & 2.79 \\
        CodeLlama-13B & 2.06 & 2.06 & 2.25 & 2.75 & 2.99 & 2.63 \\
        \bottomrule
    \end{tabular}
}
\end{table}

\begin{table}[t]
    \caption{The resource consumption comparison on RepoEval using DeepSeek-Coder-6.7B.}
    \label{tab: overhead}
    \centering
    \begin{tabular}{lccc}
        \toprule
        \textbf{Approach} & \textbf{GPU-VRAM} & \textbf{CPU} & \textbf{Memory} \\
        \midrule
        Autoregressive & 21GB & 114\% & 1.0GB \\
        REST & 21GB & 335\% & 32.6GB \\
        \ourmodel{} & 21GB & 128\% & 2.8GB\\
        \bottomrule
    \end{tabular}
    \vspace{-0.3cm}
\end{table}

\vspace{1mm}
\begin{custommdframed}
\textbf{Answer to RQ3:} \ourmodel{} can be flexibly integrated with existing correctness-focused RAG-based code generation approaches, achieving a speedup of over $2.6 \times$ while preserving their original performance on correctness. 
\end{custommdframed}
\vspace{-0.3cm}

\section{Discussion}

\subsection{Average Acceptance Length and Cost Analysis}

As average acceptance length can reflect the theoretical upper bound of achievable acceleration for retrieval-based approaches, we compare the average acceptance length of \ourmodel{} with REST, the strongest retrieval-based baseline in most cases.
The results are shown in Table \ref{tab: acclen}.
\ourmodel{} consistently exhibits a longer acceptance length across all datasets and backbone models, with an increase exceeding 50\% compared on RepoEval.
Specifically, the size of REST's datastore is approximately 8 times that of \ourmodel{}, but \ourmodel{} still achieves a higher acceleration upper bound, which demonstrates the importance of multi-source datastore as well as the context- and LLM preference-aware caching. 
Moreover, we further compare the resource consumption of \ourmodel{} with REST and autoregressive decoding, as reported in Table \ref{tab: overhead}. Since neither \ourmodel{} nor REST introduces an additional draft model, they incur no extra GPU overhead compared with autoregressive decoding. In comparison, \ourmodel{} shows markedly lower resource consumption than REST: its memory usage is less than one tenth, and its CPU usage is about one third, comparable to autoregressive decoding.
With a compact yet effective datastore and accelerated construction speed, \ourmodel{} provides a more lightweight and efficient inference acceleration approach.
In conclusion, \ourmodel{} can achieve a high average acceptance length, reflecting its ability to generate high-quality drafts, while also attaining a higher speedup ratio.

\subsection{Heatmap of \ourmodel{}'s Retrieval Performance}
\begin{figure}[t]
 \setlength{\abovecaptionskip}{0.1cm}
    \begin{subfigure}[b]{0.48\linewidth}
        \includegraphics[width=\linewidth]{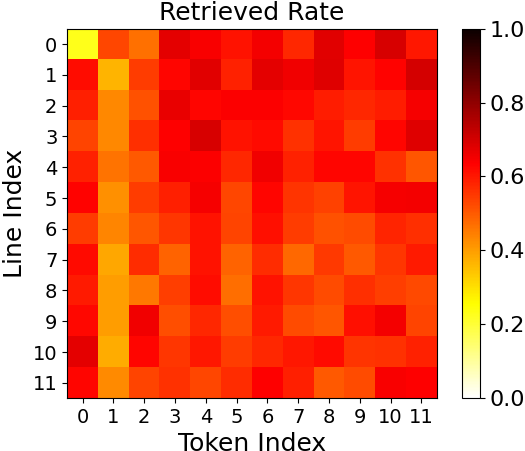}
        \caption{}
    \end{subfigure}
    \begin{subfigure}[b]{0.48\linewidth}
        \includegraphics[width=\linewidth]{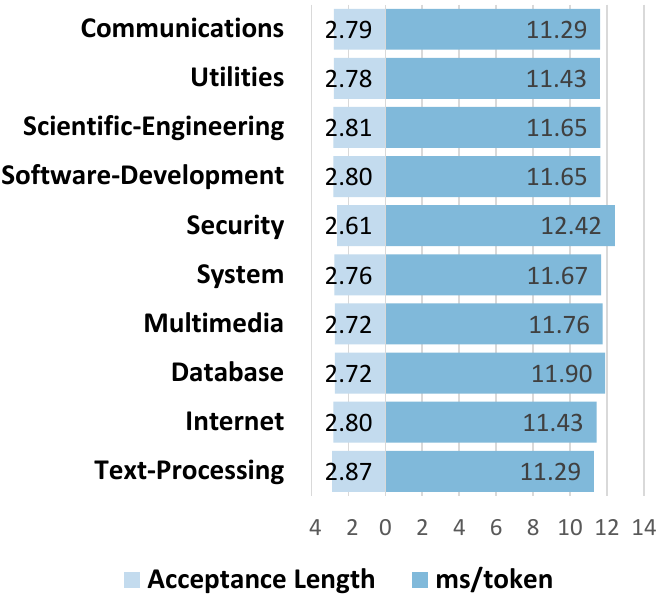}
        \caption{}
    \end{subfigure}
  \caption {(a) Retrieval performance of \ourmodel{}; (b) Performance of \ourmodel{} on different code topics. }
  \label{fig: acc comparison and heatmap}
  \vspace{-0.2cm}
\end{figure}

As mentioned in Section \ref{sec: observations}, previous retrieval-based approaches often fail at the first non-whitespace token in each code line.
To explicitly illustrate \ourmodel{}'s effectiveness, we depict its retrieval performance heatmap in Fig. \ref{fig: acc comparison and heatmap}(a), with the evaluated samples and experimental settings aligned with Fig. \ref{fig: retrieval analysis}(a).
A clear observation is that Fig. \ref{fig: acc comparison and heatmap}(a) has a markedly darker color intensity, especially at the first non-whitespace token, indicating a significant increase in the probability of \ourmodel{} retrieving valid results. 
This improvement can be attributed to two factors: on the one hand, our retrieval strategy probabilistically avoids retrievals at such positions; on the other hand, the multi-source datastore and the context- and LLM preference-aware \textsc{cache} store high-quality sequences, which enhances the retrieval hit rate.
Overall, this comparison underscores the enhanced retrieval efficacy of \ourmodel{}.

\subsection{Performance on Different Topics}

In practical software development, repositories are typically categorized into various domains, such as web development, text processing, \textit{etc}. Each domain tends to exhibit distinct coding styles and conventions. Therefore, to investigate the generalization capacity of \ourmodel{} across different domains, we report Deepseek-Coder-6.7B results on DevEval’s 10 topics. As shown in Fig. \ref{fig: acc comparison and heatmap}(b), \ourmodel{} demonstrates consistent and substantial acceleration across all topics, highlighting its robustness across diverse contexts.

\subsection{Case Study}

\begin{figure}[t]
\centering
  \includegraphics[width=0.95\linewidth]{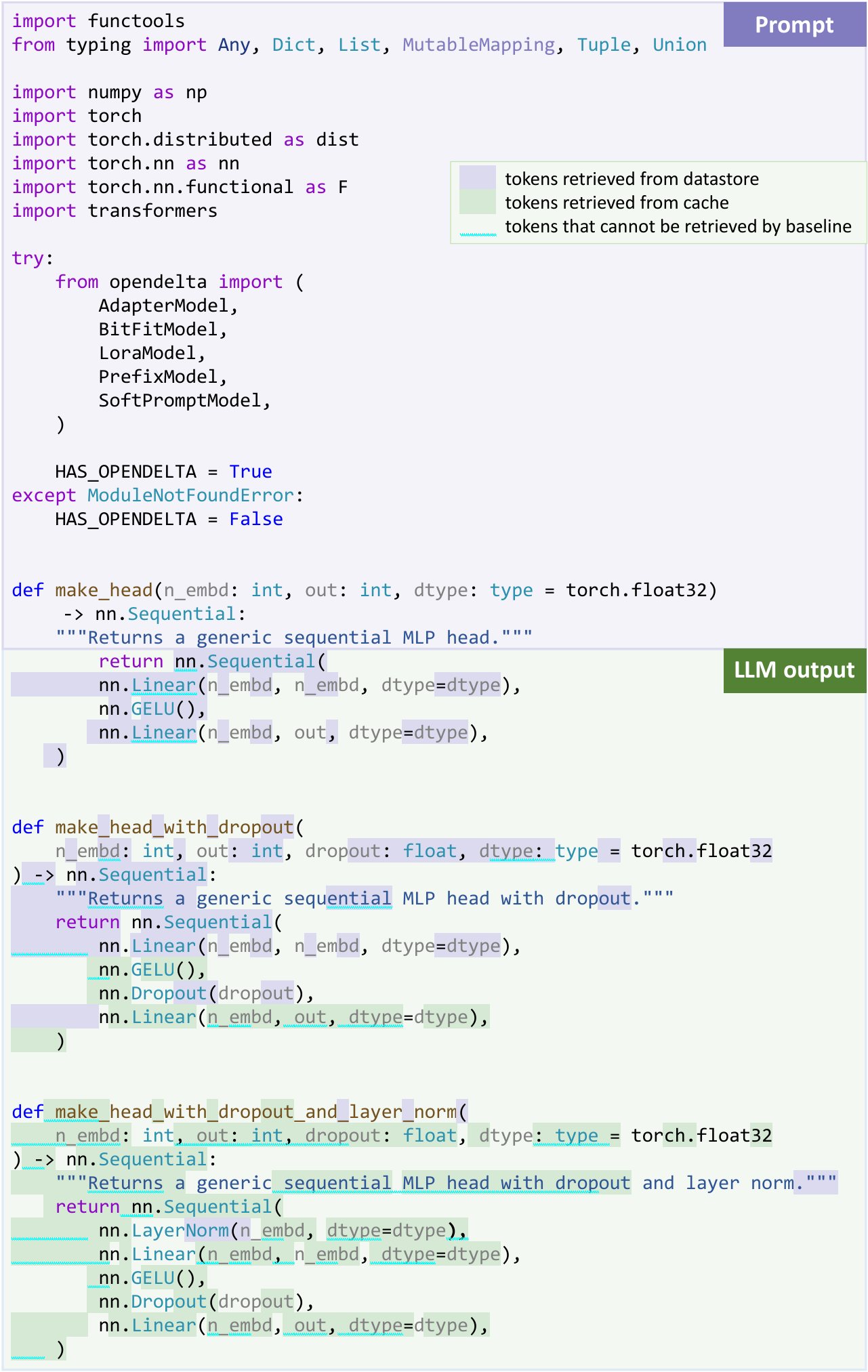}
  \caption {Case study of an example sourced from RepoEval, with legends positioned in the upper-right corner. }
  \label{fig: case}
\end{figure}

To further illustrate the effectiveness of \ourmodel{}, we conduct a case study. As shown in Fig. \ref{fig: case}, we use different background colors to highlight the sources of the accepted draft tokens utilizing \ourmodel{}.
Additionally, the underlined tokens refer to those that can be retrieved by \ourmodel{} but not by the baseline (REST with $D_c$ as the datastore).
When generating the earlier parts, the \textsc{cache} remains unavailable due to an insufficient accumulation of sequences. Nonetheless, lots of repository-related tokens can be additionally retrieved by \ourmodel{}, benefiting from the multi-source datastore.
When the \textsc{cache} is available, a larger number of consecutive tokens becomes retrievable, thereby enhancing the inference speed by extending acceptable sequences and reducing retrieval overhead.

\subsection{Threats to Validity}

\textbf{Internal.}
The internal threat to validity lies in the potential risk of data leakage during the construction of the datastore $D$, as the code contained within the datastore $D$ may overlap with the code that is intended to be generated. 
To mitigate this issue, we adopted precautionary measures during the construction of datastore $D$.
Specifically, when building the repository-related datastore $D_r$, we excluded all portions requiring generation, thereby ensuring that $D_r$ contains no ground-truth code.
For the common datastore $D_c$, where potential leakage may occur due to the use of The Stack \cite{kocetkov2022stack}, we examined the collection timelines of all relevant resources: DevEval \cite{li2024deveval} (downloaded in November 2023), RepoEval \cite{zhang2023repocoder} (repositories created after January 1, 2022), and The Stack (collected from January 2015 to June 2022). These timelines indicate that The Stack and DevEval share no temporal overlap, with only minor overlap possible with RepoEval. Moreover, only 1/145 of the Python subset of The Stack was utilized, thereby substantially reducing the likelihood of leakage. As previous studies \cite{he2024rest} have also employed The Stack to construct datastores, we adopt a consistent practice in this work.

\textbf{External.}
The external threat to validity stems from dataset and backbone model limitations, as we perform experiments on a limited set of datasets and models. Thus, the results obtained may not generalize to other experimental settings. 
To mitigate this limitation, we evaluate \ourmodel{} on three datasets spanning both repository-level and standalone code generation tasks, where \ourmodel{} consistently outperforms baseline approaches, confirming its generalization.
Furthermore, \ourmodel{} maintains robust performance regardless of the underlying LLM, validated through systematic experiments with LLMs ranging from 1B to 13B parameters.
Although we do not provide results on larger models due to resource constraints, \ourmodel{}'s LLM preference-aware cache helps ensure that retrieved drafts remain effective by aligning them with the output style of the target LLM, thereby supporting generalization across different model scales.

\section{Conclusion}
In this paper, we propose \ourmodel{}, a simple yet efficient LLM inference acceleration approach for repository-level code generation without compromising generation quality.
\ourmodel{} leverages a multi-source datastore as well as a context- and LLM preference- aware cache to improve the acceptance length of the retrieved draft, while minimizing redundant retrieval operations through a dynamic and efficient retrieval strategy.
Experimental results demonstrate that \ourmodel{} outperforms state-of-the-art inference acceleration approaches on both repository-level and standalone code generation tasks.
\ourmodel{} can also be integrated with correctness-focused code generation approaches to accelerate their inference speed.
Requiring no draft model or additional training, \ourmodel{} provides a lightweight and practical solution for LLM inference acceleration in code generation.

\section*{Acknowledgment}
This research is supported by the National Natural Science Foundation of China Grants Nos. 62302021 and 62332001.
We thank Borui Zhang and Runlin Guo for their contributions to the demo development.

{\small
\bibliographystyle{IEEEtran}
\bibliography{references}
}

\end{document}